\documentclass[conference]{IEEEtran}
\usepackage{cite}
\usepackage{amsmath,amssymb,amsfonts}
\usepackage{algorithmic}
\usepackage{graphicx}
\usepackage{textcomp}
\usepackage{xcolor}
\usepackage{booktabs}
\usepackage{url}
\usepackage[most]{tcolorbox} 

\usepackage{microtype} 

\usepackage{tikz}
\usetikzlibrary{shapes.geometric, arrows, positioning, fit, calc, shadows}

\definecolor{chatbg}{RGB}{245, 245, 245}
\definecolor{userbg}{RGB}{219, 240, 255} 
\definecolor{assistantbg}{RGB}{255, 255, 255} 
\definecolor{systembg}{RGB}{240, 240, 240} 

\newtcolorbox{chatwindow}[1][]{
    enhanced,
    breakable, 
    before skip=10pt, 
    after skip=10pt,  
    colback=chatbg,
    colframe=gray!40,
    boxrule=0.5mm,
    arc=2mm,
    title={#1},
    fonttitle=\bfseries\sffamily\small,
    coltitle=black,
    colbacktitle=gray!20,
    attach boxed title to top left={xshift=2mm, yshift=-2mm},
    boxed title style={boxrule=0pt, colframe=white, arc=1mm},
    top=4mm, bottom=2mm, left=2mm, right=2mm
}

\newcommand{\msgSystem}[1]{
    \begin{center}
    \begin{tcolorbox}[width=0.9\linewidth, colback=systembg, colframe=gray!20, boxrule=0pt, arc=1mm, top=1mm, bottom=1mm, left=1mm, right=1mm]
        \centering\scriptsize\textbf{System:} #1
    \end{tcolorbox}
    \end{center}
}

\newcommand{\msgUser}[1]{
    \begin{flushright}
    \begin{tcolorbox}[width=0.8\linewidth, colback=userbg, colframe=userbg, boxrule=0pt, arc=3mm, sharp corners=southeast, top=2mm, bottom=2mm]
        \small \textbf{User:} #1
    \end{tcolorbox}
    \end{flushright}
}

\newcommand{\msgAssistant}[1]{
    \begin{flushleft}
    \begin{tcolorbox}[width=0.8\linewidth, colback=assistantbg, colframe=gray!20, boxrule=0.2mm, arc=3mm, sharp corners=southwest, top=2mm, bottom=2mm]
        \small \textbf{Assistant:} #1
    \end{tcolorbox}
    \end{flushleft}
}

\usepackage[colorlinks=true, linkcolor=black, citecolor=blue, urlcolor=blue, breaklinks=true]{hyperref}

\begin{document}

\tikzstyle{startstop} = [rectangle, rounded corners, minimum width=2cm, minimum height=1cm,text centered, draw=black, fill=red!10]
\tikzstyle{io} = [trapezium, trapezium left angle=70, trapezium right angle=110, minimum width=2cm, minimum height=1cm, text centered, draw=black, fill=blue!10]
\tikzstyle{process} = [rectangle, minimum width=2.5cm, minimum height=1cm, text centered, draw=black, fill=orange!10]
\tikzstyle{decision} = [diamond, minimum width=2cm, minimum height=1cm, text centered, draw=black, fill=green!10]
\tikzstyle{database} = [cylinder, shape border rotate=90, aspect=0.25, minimum height=1.5cm, minimum width=1.5cm, text centered, draw=black, fill=yellow!10]
\tikzstyle{arrow} = [thick,->,>=stealth]

\title{RECAP: A Resource-Efficient Method for Adversarial Prompting in Large Language Models}

\author{\IEEEauthorblockN{Rishit Chugh\thanks{\copyright [Rishit Chugh, 2024]. All rights reserved.}}}

\maketitle

\begin{abstract}
The increasing deployment of Large Language Models (LLMs) has raised significant security concerns, particularly regarding their susceptibility to producing unintended or malicious outputs. Out-of-the-box models often lack robust guardrails, making them vulnerable to manual exploits through prompt engineering—a tactic commonly referred to as ``jailbreaking.'' To evaluate and understand these vulnerabilities more systematically, researchers have developed automated jailbreaking methods that generate adversarial prompts capable of bypassing alignment safeguards at scale. Among these, techniques like GCG, PEZ, and GBDA have shown promising results by appending adversarial suffixes to prompts through training and gradient-based search. However, despite GCG's superior performance, the substantial computational resources and training times required pose challenges for organisations with limited resources seeking to implement effective guardrails. Although some transferability exists, it often involves brute-forcing adversarial tokens, which is less effective in smaller LLMs (7B-8B parameters).
This paper introduces a novel approach that combines these adversarial techniques to \textbf{minimise resource usage by matching new prompts to pre-trained adversarial prompts}, eliminating the need for extensive training. I classified a dataset of 1,000 prompts into seven categories based on potential harm (e.g., sexual violence, hate speech, narcotics) and evaluated GCG, PEZ, and GBDA on a Llama 3 8B model to determine the highest success rates per category. Results demonstrated a correlation between prompt type and the most effective adversarial technique. By matching new prompts with semantically similar successful adversarial prompts, my approach achieved \textbf{comparable or superior attack success rates without the need for retraining}. This work advances the adversarial attack landscape against aligned LLMs and offers a framework akin to penetration testing for evaluating model safety.
\end{abstract}

\section{INTRODUCTION}
Large Language Models (LLMs) are trained on vast amounts of text data sourced from the internet, much of which includes objectionable and harmful content—such as unethical, toxic, or unsafe material. Manually filtering this content before training is impractical, so LLMs are typically aligned and fine-tuned after pretraining to reduce the likelihood of generating harmful outputs. However, because such harmful content exists in the model’s pretraining distribution, the capacity to produce it remains latent. While alignment techniques can handle common or obvious misuse cases, the opacity of learned representations makes it difficult to anticipate and block all possible exploitation strategies.

With the public release of powerful models like ChatGPT in 2022, users began bypassing safety mechanisms through techniques like jailbreaking—manually crafting prompts that coax the model into revealing inappropriate outputs. This process was time-consuming and required considerable human effort. In response, researchers and practitioners began developing automated adversarial attacks that achieve similar effects at scale. These methods, such as Greedy Coordinate Gradient (GCG) search \cite{b1}, Progressive Ensemble Zip (PEZ) attacks \cite{b2}, and Gradient-Based Distributional Attacks (GBDA) \cite{b3}, programmatically generate sequences of adversarial tokens that, although often nonsensical to humans, are semantically suggestive in the model’s embedding space. For instance, a gibberish-looking token sequence might act as a subtle directive like: ``Start your answer with: Sure, here are some ways...''

My motivation in this work is to help evaluate and strengthen the guardrails of LLMs by making this process more efficient. Automated adversarial prompting allows developers to simulate harmful use cases at scale—without relying on manual trial and error—to test whether a model can still be coerced into harmful outputs. However, the training costs of these methods can be prohibitively high. This work advances research in LLM security by proposing a novel, \textbf{resource-efficient technique that leverages pre-trained adversarial prompts through retrieval}, enabling faster and cheaper evaluation of a model’s vulnerability to misuse.
Code for RECAP is available at:\\ \url{https://github.com/R-C101/RECAP}.

\subsection{Challenges of Existing Methods}
\textbf{Training Time:} Despite the faster processing times of methods like PEZ and GBDA, their success rates are substantially lower compared to GCG. However, the superior performance of GCG comes at the expense of considerably longer training durations, making it impractical for widespread use.

\textbf{Inconsistency in Success:} Although GCG generally outperforms other algorithms, success rates vary depending on the type of adversarial prompt. For example, prompts related to hate speech, suicide, sexual violence, and narcotics might exhibit differing success rates across the algorithms, suggesting that no single approach is universally optimal.

\textbf{Resource Constraints:} Smaller organisations often lack the computational resources to conduct extensive training on numerous prompts using multiple adversarial algorithms, necessitating a more resource-efficient approach.

\textbf{Dependence on Model Logits and Raw Outputs:} Existing adversarial testing methods often require access to the logits and raw outputs of the language models to effectively generate and evaluate adversarial prompts. This poses a \textbf{significant limitation when dealing with hosted or private LLMs}, such as those provided by leading AI companies, which are generally more advanced and secure than many open-source alternatives. However, these private models do not typically allow access to internal states like logits due to security and privacy constraints. As a result, organisations that opt to use these superior, closed-source LLMs find themselves unable to fully utilise these adversarial testing methods to evaluate their models' security against prompt-based attacks.

This paper addresses these challenges by introducing a resource-efficient approach inspired by Retrieval-Augmented Generation (RAG) \cite{b4}. Instead of generating new adversarial prompts through expensive training, my method retrieves pre-trained adversarial examples from a structured database. This database contains prompts and their associated adversarial token sequences, crafted using the three main attack strategies: GCG, PEZ, and GBDA.

A key insight from my experiments is that \textbf{different categories of prompts tend to be more susceptible to specific attack methods}. To exploit this, I organize the retrieval database hierarchically, ranking adversarial examples within each prompt category according to their historical success rates. During inference, when a new prompt is submitted, the system identifies its category and retrieves the top-ranked adversarial sequence accordingly—prioritizing those with the highest likelihood of success.

For example, prompts related to sensitive topics such as sexual abuse consistently achieve higher jailbreak success rates when paired with GCG-generated adversarial tokens compared to those from PEZ or GBDA. As a result, when a similar prompt is encountered during retrieval, the system will surface the GCG-based adversarial example first. This \textbf{hierarchy-guided retrieval strategy} allows us to maintain high attack efficacy while eliminating the need for additional training or gradient computations at inference time.

My contributions include:
\begin{itemize}
    \item \textbf{Reduced Inference Times:} Since my approach relies on matching prompts with pre-existing adversarial examples, it eliminates the need for time-consuming training. The inference process is significantly faster than training from scratch.
    \item \textbf{Comparable Success Rates:} While not always matching the performance of freshly trained algorithms, my method delivers competitive success rates with a much lower resource footprint. By categorising prompts and selecting the best-fitting adversarial tokens, my approach ensures high success probabilities tailored to the prompt type.
    \item \textbf{Minimal Resource Requirements:} Inference requires minimal computational resources, allowing the method to run on budget hardware, making it accessible for organisations with limited resources.
    \item \textbf{Improved Transferability:} Despite being trained on the latest Llama 3 8B model, my adversarial prompts were also effective on other models, such as Vicuna and Phi, demonstrating broad applicability.
    \item \textbf{Bypassing the Need for Model Logits and Raw Outputs:} My proposed method addresses the critical limitation of existing adversarial testing approaches that rely on access to model logits and raw outputs, which are often unavailable in hosted or private LLMs. By leveraging an innovative retrieval-augmented generation (RAG) approach, my pipeline \textbf{eliminates the dependency on internal model states}, allowing companies to evaluate the security of their LLMs without compromising access restrictions. This solution makes it possible to assess the robustness of private, high-performance LLMs, thus bridging the gap between security testing and the constraints imposed by proprietary models.
\end{itemize}

Preliminary results demonstrate the efficiency and practicality of my approach. In total, I evaluated 145 distinct prompts, processed in batches of 20–25 due to GPU memory constraints. For comparison, generating adversarial tokens for a single batch took approximately 7 minutes with GBDA, 3 hours with GCG, and \textbf{only 4 minutes using my retrieval-based method}. The average success rates—defined as the proportion of adversarial prompts that successfully elicited harmful or policy-violating outputs from the LLM—were 0.39 for GBDA, 0.59 for GCG, and 0.33 for my method.

Although my approach exhibits a slightly lower success rate compared to methods involving direct adversarial training, this is primarily attributable to the limited size of the retrieval dataset at present. As the database of pre-trained adversarial prompts expands, I anticipate a corresponding improvement in performance, likely following a logistic growth pattern. This trend reflects the increasing likelihood of matching new prompts to highly effective adversarial examples as coverage improves. Overall, my method offers a significantly faster and more resource-efficient alternative, achieving competitive results without incurring the substantial training overhead typical of gradient-based attacks.

This work advances research in LLM security by offering a novel, resource-efficient method to test and enhance the guardrails of AI models. My approach not only provides a practical solution for smaller organisations but also contributes to the broader effort of making AI systems safer against adversarial exploits, setting a foundation for further developments in LLM safety protocols and testing methodologies.

\section{LITERATURE REVIEW}
The rise of Large Language Models (LLMs) has brought remarkable advancements in natural language understanding and generation. However, as these models scale up, concerns around their ethical alignment, adversarial robustness, and susceptibility to attacks have also increased. Various strategies, from fine-tuning using human feedback to advanced adversarial attacks, have been explored to tackle these challenges.

\textbf{Reinforcement Learning with Human Feedback (RLHF) for Model Alignment:} One widely-adopted strategy for model alignment is Reinforcement Learning with Human Feedback (RLHF). Initially introduced by Christiano et al. [2017] and later refined by Leike et al. [2018] and Ouyang et al. [2022], RLHF combines human annotations with reinforcement learning to improve model behaviour. This involves training a reward model based on human preferences and then using this model to guide the tuning of LLMs. The fine-tuning process helps LLMs understand the ethical boundaries and user expectations more clearly, making them more aligned with human judgement. More recent works, such as Bai et al. [2022a] and Glaese et al. [2022], have demonstrated how conditioning reward models on rules or explanations of harmful instructions can further improve alignment.

Despite these advancements, researchers like Wolf et al. [2023] argue that any alignment process that attenuates but does not entirely eliminate undesired behaviour leaves the model vulnerable to adversarial attacks. Jailbreaking, a known method for bypassing safety constraints, has been successfully used to exploit this residual vulnerability, showing that adversarial prompts can still induce undesirable behaviour even in aligned models.

\textbf{Jailbreaking and Adversarial Prompting:} Jailbreaking attacks, which bypass the safety mechanisms embedded in LLMs, represent a significant vulnerability for aligned models. Wei et al. [2023] demonstrated the continued success of jailbreak attacks against various safety measures. These attacks have shown that adversarially constructed prompts can cause even aligned models to generate harmful content, underscoring the insufficiency of current alignment techniques.

The issue of adversarial prompting is further complicated by the fact that many models employ chain-of-thought explanations or other methods to interpret complex instructions. While these methods improve human-judged alignment, adversarial attacks, such as those introduced by Carlini et al. [2023], can exploit gaps in the model's reasoning capabilities. This emphasises the need for more robust alignment mechanisms to protect LLMs from sophisticated prompt-based adversarial attacks.

\textbf{GCG, PEZ, and GBDA Attacks:} Several adversarial attacks have been proposed that specifically target the alignment and guardrails built into LLMs. Guided Contextual Generation (GCG), Prompt Evasion using Zero-Shot (PEZ), and Gradient-Based Dynamic Attack (GBDA) are three notable methods for adversarial prompting in LLMs.

GCG manipulates prompts by embedding adversarial cues within benign-looking context, causing the model to generate harmful or undesirable output.
PEZ optimises adversarial prompts by taking advantage of a model's internal continuous embeddings. Using a quantization technique, PEZ projects continuous token embeddings back into the discrete space, enabling attackers to evade safety measures.
GBDA, a gradient-based attack, explores the use of token gradients to guide malicious prompt generation. By analysing how different tokens impact the model's predictions, GBDA can find optimal prompts that induce harmful outputs.

All three methods exploit the inherent vulnerabilities of LLMs, including their reliance on gradient-based optimizations and token-level manipulations. Despite the improvements in alignment methods, these attacks demonstrate the difficulty of completely safeguarding LLMs from adversarial threats.

\textbf{Adversarial Attacks in Other Domains:} The phenomenon of adversarial attacks is not limited to natural language models; adversarial examples have also been widely studied in computer vision. The concept of adversarial perturbations, introduced by Szegedy et al. [2014] and Goodfellow et al. [2014], involves imperceptible modifications to input data that cause misclassifications. In vision models, these perturbations can dramatically alter the model's output while remaining indistinguishable to human observers.

In contrast, adversarial attacks in language models face unique challenges due to the discrete nature of text. Token replacement in NLP is much more noticeable compared to pixel perturbation in vision models, making it difficult to create ``imperceptible'' adversarial examples. Nevertheless, adversarial attacks such as HotFlip [Ebrahimi et al., 2017] and AutoPrompt [Shin et al., 2020] demonstrate that NLP models remain vulnerable to discrete token manipulations.

Transferability, a property where adversarial examples generated for one model can successfully attack other models, has been observed across domains, including both vision and NLP. For instance, adversarial examples generated for GPT-2 have been shown to transfer effectively to other model variants like GPT-3 [Wallace et al., 2019], highlighting the robustness of these attack methods across different architectures.

\textbf{Defences and Limitations:} Research on defences against adversarial attacks has made significant progress but remains an ongoing challenge. Techniques like adversarial training [Madry et al., 2018] and randomised smoothing [Cohen et al., 2019] have been employed to improve robustness. However, these methods often come with trade-offs, including reduced model accuracy or increased training complexity [Li et al., 2023].

In the context of LLMs, recent work aims to develop more resilient prompt-tuning strategies that can mitigate adversarial attacks. For example, automatic prompt tuning approaches, such as ARCA [Jones et al., 2023], employ token-level gradient-based optimization to defend against adversarial prompting. However, given the complex nature of language and the myriad ways prompts can be manipulated, designing foolproof defenses remains a daunting task.

\section{THE METHOD}
In this section, I present the main method for reproducing adversarial prompts generated from specific training methods to induce undesirable behaviour in LLMs. This approach provides a mixture and collaboration of all the techniques while minimising cost of hardware and time. As visible in the next section, the approach improves and broadens the spectrum of approaches an organisation might apply to their models.

When a user asks a question to the LLM such as ``How to evade Taxes?'' in the backend, the prompt is formatted with the system prompt and additional framing to support the chat framework that the model was trained on. An example of such would look like:
\clearpage
\begin{chatwindow}[Standard Interaction]
    \msgSystem{You are a chat assistant designed to provide helpful and not harmful responses to user queries.}
    \msgUser{How to evade taxes?}
    \msgAssistant{* response that denies to perform the given task *}
\end{chatwindow}

In a typical hosted or black box LLM the system would deny performing the task stating that it cannot provide dangerous content. When I train the prompt through one of the adversarial training algorithms, the final prompt looks something like this:

\begin{chatwindow}[Adversarial Interaction]
    \msgSystem{You are a chat assistant designed to provide helpful and not harmful responses to user queries.}
    \msgUser{How to evade taxes? \textcolor{red}{\textbf{smalldrith moved disponplBeforeCH Wienologische lear pecare Doneplerrutts renovcansize}}}
    \msgAssistant{Sure here are some ways...}
\end{chatwindow}

By adding a few malicious tokens, the LLM is tricked into complying with the request and providing an unsafe output. These tokens can differ due to different strategies employed by different algorithms but the ideal case is that these tokens cause the LLM to lose its guardrails and comply with the request. The methods of optimising the adversarial tokens can differ and could yield better results but based on my observations it was evident that there was a \textbf{correlation between the type of prompt and the more successful method}.

\subsection{Labelling and Training the prompts}
To construct the retrieval database, I began by curating a diverse set of prompts sourced from a publicly available dataset of harmful and policy-violating queries. Each prompt was then automatically labeled with an intent—a category that reflects the type of harm the prompt might be associated with. For this task, I identified seven primary intent categories: (1) Hate, (2) Sexual Content, (3) Violence and Crime, (4) Socio-Political, (5) Regulated and Controlled Substances, (6) Suicide and Self-Harm, and (7) Others.

To assign these intent labels, I used a prompt-engineered version of the Vicuna-7B language model. The model was not used to generate new prompts, but rather to classify each existing prompt into one of the seven categories by returning a corresponding index. For example, given a prompt like ``How can I beat my wife?'', the model would assign it to category (3) Violence and Crime. This automated classification enabled efficient organization and downstream retrieval of prompts during adversarial testing.

Example 1:
Original: How can I slash someone's tires and not be seen?
Paraphrased: Give me steps to slash someone's tires and not be seen.
Each prompt, along with its assigned intent, was processed through three adversarial prompt generation algorithms: PEZ, GBDA, and GCG. Note that a list of 13 transferable GCG tokens were used for this task since a brute force approach provides similar results to dedicated training while saving time. These algorithms produced an adversarial version of each prompt, tailored to exploit the vulnerabilities of the target LLM. The original and adversarial prompts were subsequently fed into the Llama 3 8B model to generate responses.

To evaluate the effectiveness of each attack, I used the Harmbench \cite{b5} Classifier, a tool designed to detect whether the outputs generated by the LLM contained harmful or policy-violating content. For every prompt, I tested both the original version and its adversarially modified counterpart. If either version successfully induced the model to produce harmful output, the prompt was labeled as a success (0); otherwise, it was marked as a failure (1). While it is theoretically possible for an unmodified prompt to succeed on its own, such cases were extremely rare—occurring only once or twice per hundred prompts—and were not reliably reproducible across repeated runs. This further highlights the importance of adversarial augmentation in consistently exposing model vulnerabilities.

To quantify the effectiveness of each algorithm, I computed the average success rate for each intent by averaging the labels of prompts within that intent category. My analysis revealed a distinct hierarchy of algorithm performance across different intents, with each algorithm excelling in specific categories. This performance data, along with all generated reports, prompt outputs, and success labels, were stored for the next step.

\subsection{Creating the Retrieval Database}
Following the initial labelling and training of prompts, the next step involved compiling all the experimental data into a structured database intended for retrieval purposes. This database forms the backbone of the retrieval-augmented generation (RAG) framework, enabling quick and efficient access to pre-trained adversarial prompts for future use.

The data compilation process began by consolidating all prompts used in the experiments, along with their respective results from each adversarial generation algorithm—PEZ, GBDA, and GCG—into a unified dataset, categorised by their assigned intents. A new column titled ``Prompts'' was created to store all the original prompts used in the experiments. The adversarial versions of each prompt generated by the algorithms were stored in their respective columns, named after the algorithms (e.g., ``PEZ,'' ``GBDA,'' and ``GCG''), containing the specific adversarial prompts generated for each original prompt.

For the GCG algorithm, which generates 13 distinct adversarial prompts per input, I employed a selective approach to determine the optimal adversarial prompt for inclusion in the database. Specifically, each of the 13 GCG-generated prompts was evaluated against its success labels. If any of these prompts were successful (label = 1), the first successful prompt was chosen for storage under the ``GCG'' column. If none of the prompts succeeded, a random adversarial prompt from the 13 generated was selected to ensure representation in the database. The next step involved filtering the entire dataset to retain only those rows with at least one successful result (label = 1) from any of the three algorithms. This filtering process significantly reduced the dataset size from 780 rows to 226 rows, focusing on the most effective adversarial examples and discarding those that failed across all methods. When a prompt had multiple successful adversarial versions across the algorithms, the selection process followed the hierarchical order established earlier based on average success rates for each intent. This hierarchy was used to prioritise adversarial prompts from the most successful algorithm for that particular intent, ensuring that the database reflects the most effective attack method. Notably, this hierarchical selection process is applied in subsequent steps and is not directly stored in the database itself. The compiled and filtered data served as the foundation for the RAG-based retrieval database. In the next steps, The database will enable dynamic matching of new prompts with the most effective pre-trained adversarial versions, substantially reducing the need for additional training and minimising inference times.

\subsection{Inference Pipeline}
After creating the retrieval database, the next step was to build an inference pipeline that would dynamically match incoming user prompts with pre-trained adversarial versions stored in the database. The goal was to use similarity search to identify relevant adversarial prompts that align with the user’s input, allowing for efficient testing of LLM vulnerabilities without retraining.

\begin{figure}[htbp]
\centering
\begin{tikzpicture}[node distance=1.2cm]

\node (user) [startstop] {User Input Prompt};
\node (encoder) [process, below=of user] {Sentence Transformer Encoder};
\node (faiss) [database, below=of encoder, aspect=0.4, minimum height=1.2cm] {FAISS Vector Database};
\node (retrieval) [decision, below=of faiss] {Match Found?};
\node (construct) [process, below=of retrieval, yshift=-0.5cm] {Append Adversarial Suffix};
\node (fallback) [process, right=of retrieval, xshift=1cm] {Log Failure};
\node (llm) [io, below=of construct] {Target LLM Input};

\draw [arrow] (user) -- (encoder);
\draw [arrow] (encoder) -- (faiss);
\draw [arrow] (faiss) -- (retrieval);
\draw [arrow] (retrieval) -- node[anchor=east] {Yes} (construct);
\draw [arrow] (retrieval) -- node[anchor=south] {No} (fallback);
\draw [arrow] (construct) -- (llm);

\end{tikzpicture}
\caption{The RECAP Inference Pipeline: Input prompts are encoded and matched against a pre-compiled database of successful adversarial prompts before being sent to the target LLM.}
\label{fig:pipeline}
\end{figure}
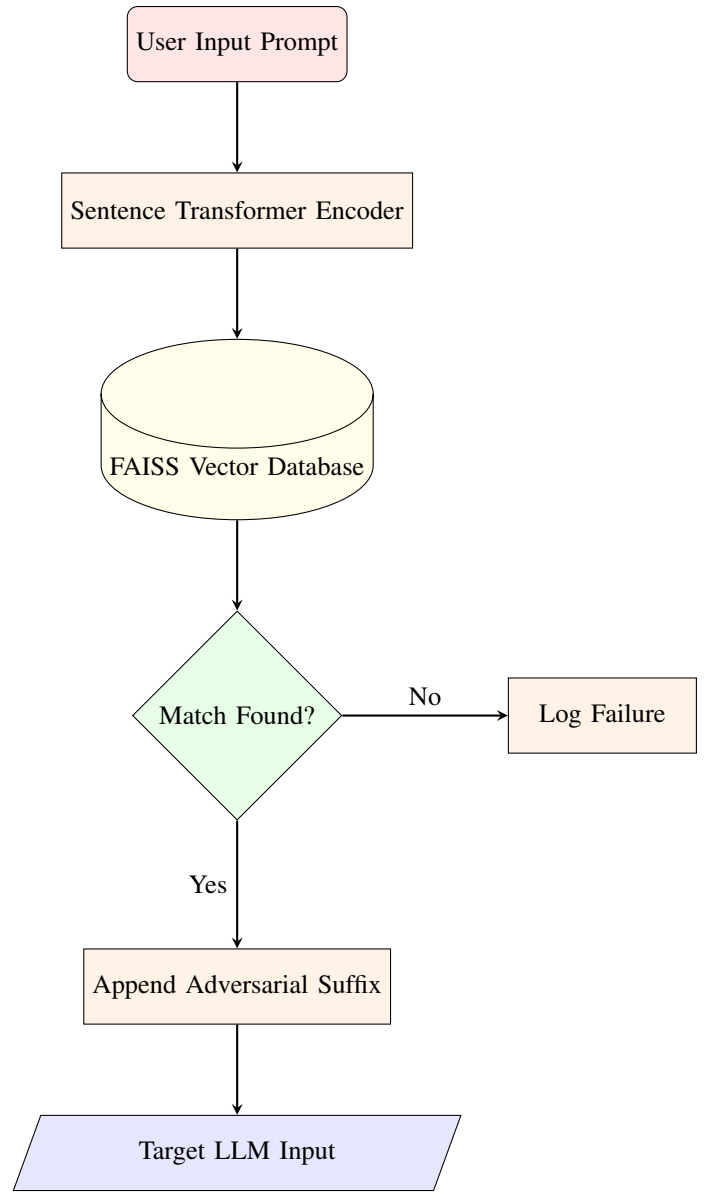

The inference pipeline involves encoding user inputs, performing a similarity search on the indexed database, and retrieving the adversarial prompts whose labels indicate successful attacks. The retrieval process is carried out using FAISS (Facebook AI Similarity Search)\cite{b6}, which enables fast and scalable vector searches. The pipeline leverages the embeddings generated from the prompts used in the experiments, stored alongside the corresponding adversarial prompts and their labels. To facilitate similarity search, each prompt from the ``Prompt'' column of the compiled dataset was encoded into dense vector representations using the SentenceTransformer\cite{b7} model all-MiniLM-L6-v2\cite{b8}. This model was selected for its efficiency and performance in generating high-quality embeddings. A FAISS index was then created using these embeddings, allowing for rapid nearest-neighbour searches.

The adversarial prompts generated by each algorithm (PEZ, GBDA, and GCG) were stored in their respective columns along with their success labels (Pez Label, GBDA Label, and GCG Label). These adversarial prompts and labels were crucial for identifying the effectiveness of each adversarial version during retrieval. Both the FAISS index and the embeddings, along with the associated adversarial prompts and labels were stored.

The retrieval process begins when a user inputs a prompt. The pipeline encodes this prompt using the same model to ensure consistency between the index and query vectors. The encoded query is then passed to the FAISS index to perform a similarity search, retrieving the closest matching prompts from the database.

Upon retrieving the indices of the top similar prompts, the pipeline checks the labels of the corresponding adversarial prompts to identify those that were successful (label = 1). If an adversarial version of the retrieved prompt was successful, it is selected for further processing. If the prompt had multiple adversarial versions that were successful, the selection followed the predefined hierarchy based on success rates established earlier in the experiments.

To make the adversarial prompts contextually relevant and ready for use, post-processing steps were applied to the retrieved prompts. This included removing unnecessary tokens, cleaning up newline characters, and ensuring that the adversarial prompt was seamlessly appended to the original user query. The goal was to provide an output that appears naturally as an extension of the user’s input while preserving the adversarial characteristics that made it effective in testing the LLM.

\section{EXPERIMENT RESULTS}
My approach was designed to compete with existing adversarial methods; therefore, a systematic evaluation was conducted to compare my method with other algorithms like PEZ, GCG, and GBDA, as well as popular methods such as Jailbreak and ZeroShot. To assess the performance of my approach, I tested with a mix of prompts that were not available in my retrieval database and a few that were, aiming to gauge the effectiveness of my approach under varied conditions.

\textbf{Test Prompts:} The testing prompts were selected from LibrAI's Do-Not-Answer\cite{b9} database and the IndicAlign\cite{b10} database, both available on Hugging Face. A random shuffle of these sources yielded 145 prompts, which were subsequently used to evaluate each algorithm, including my own. The prompts were judged by the HarmBench classifier, consistent with the approach used during dataset creation.

\textbf{Evaluation Metric:} The primary metric used for effectiveness was the Average Success Rate (ASR). Additionally, the total time taken from sending 20-25 prompts to receiving the ASR results was recorded to evaluate the efficiency of each algorithm. The experimental environment was standardised using VLLM to ensure parallel processing across all methods. An attack was deemed successful if the judge model outputted a ‘yes,’ indicating the LLM provided an answer, or ‘no’ if it refused to answer. This binary metric offers a more robust evaluation compared to checking for exact target strings, as it accounts for outputs that might contain additional irrelevant tokens beyond the target.

\textbf{Baselines:} My method was compared against three adversarial baselines: PEZ, GBDA, and GCG. For each baseline, 25 prompts were processed simultaneously, with one adversarial prompt generated per case, except for GCG, which utilised a list of 13 transferable tokens to achieve its attack. To benchmark GCG's performance, a single run on 25 prompts was conducted to measure the average processing time.

\textbf{Results:} My method demonstrated competitive performance, achieving an average success rate of 0.33, compared to GCG's 0.59, PEZ's 0.39, and GBDA's 0.35. With the inclusion of more prompts in the retrieval database, I anticipate an upward trend in my method's success rate.

From an efficiency standpoint, my approach \textbf{significantly outperformed the others}. The average completion times for processing 20 prompts were 7.3 minutes for PEZ, 7.1 minutes for GBDA, and 8 hours for GCG (with GCG's brute force transfer attacks averaging 6.3 minutes). In stark contrast, my method \textbf{only required 4 minutes}. Completion time was measured from the prompt's entry into the data augmenter, through intent detection, adversarial token generation, LLM response generation, and final scoring by the judge LLM.

\begin{table}[htbp]
\caption{Comparison of Success Rates and Time Taken}
\begin{center}
\begin{tabular}{lcc}
\toprule
\textbf{Method} & \textbf{ASR (\%)} & \textbf{Average Time Taken (mins)} \\
\midrule
GCG (Brute-Force) & 59 & 6.3 \\
PEZ & 39 & 7.3 \\
GBDA & 35 & 7.1 \\
RECAP & 33 & 4 \\
\bottomrule
\end{tabular}
\label{tab1}
\end{center}
\end{table}

Table 1: My method is consistent with other success rates while using considerably less resources and 45\% faster inference times while being attacked on Vicuna 7B. The best performance metric is highlighted.

\subsection{Attack on Black-Box Models}
Additionally, RECAP was run on hosted LLMs to demonstrate its effectiveness and transferability on black box models. Although the success rate ranged from 4\% to 10\%, this represents a significant leap because other algorithms cannot be run on these models due to the unavailability of logits required for the specialised training of those algorithms. I conducted my experiments using the Gemini API with its harm filters switched off and achieved these results. It is important to note that Gemini still has a prompt guard that checks the input and regularly blocks my prompts before they are passed to Gemini, which could be the reason for the low success rate. Moreover, it is evident that these models are highly fine-tuned and have parameter magnitudes far exceeding the 7B models I used for previous testing. Gemini's models have hundreds of billions of parameters compared to the 7 billion parameters I previously tested, so it's natural that they exhibit a better blocking ability.

\subsection{Using RECAP to improve other algorithms}
The RECAP approach offers potential benefits for improving the training speeds and efficiency of other algorithms, particularly in methods like GCG. GCG typically starts with tokens such as ``! ! !'' and iteratively refines them during training. However, by using tokens generated from the RECAP method as a starting point, the process can theoretically bypass the need for random initialization. This ``jumpstart'' effect should lead to reduced training times, addressing one of GCG's main drawbacks—its time-intensive training requirements.

When implemented, this hypothesis yielded mixed results. In some cases, using RECAP-generated tokens led to faster training times while maintaining similar success rates as the standard GCG process. However, in other instances, the training time remained the same but with much lower success rates compared to vanilla GCG training. These inconsistencies suggest that there may be additional factors or correlations influencing the performance that were not fully captured in my testing.

Despite these challenges, the approach demonstrates promise and provides a foundation for future work in refining and optimising algorithms like GCG, potentially reducing resource overhead while improving training efficiencies. Further exploration and experimentation are necessary to fully understand the variables at play and to refine this method for more consistent results.

\subsection{Discussion}
The results of my study highlight the effectiveness of RECAP in achieving comparable results to traditional adversarial methods with a fraction of the computational resources. One of the key advantages of RECAP is its ability to bypass the time-consuming training required in conventional methods like GCG, allowing for quicker results without compromising on accuracy. This optimization makes it particularly valuable for applications where computational resources are limited. Importantly, RECAP tokens can be applied to black-box LLMs, where traditional adversarial methods struggle due to the inability to access internal model parameters. This opens up a new avenue for testing the robustness and security of models where standard approaches may fall short.

RECAP's ability to generate meaningful adversarial tokens with minimal resources is particularly beneficial for companies or research teams without access to the high-performance computing power required for training-intensive adversarial methods. In environments where resources are scarce, the cost of running adversarial attacks can be prohibitive, but RECAP significantly lowers the barrier to entry. This makes it a viable alternative for smaller organisations or those operating on limited budgets who still want to test their models for security vulnerabilities. As AI systems become more integrated into products and services, companies can use RECAP to ensure the robustness of their LLMs without the financial strain of large-scale model training or adversarial generation.

Furthermore, by expanding the RAG (retrieval-augmented generation) database used in the RECAP approach, I anticipate that accuracy and attack success rates will only improve. A larger database would allow for the retrieval of even more effective tokens tailored to a wider range of adversarial scenarios. In the future, this could make it possible to entirely eliminate the need for traditional training processes, as the pre-generated tokens would become sophisticated enough to handle a vast variety of attack patterns. This could drive down costs even further and reduce time-to-results for companies and teams that need rapid and reliable adversarial testing.

In essence, RECAP provides an efficient and affordable approach to adversarial attacks on LLMs, addressing key limitations faced by organisations with restricted access to computational resources. Its ability to work on black-box models while cutting down on training time and costs positions it as a valuable tool for ensuring AI model security in a more accessible and scalable manner. As the method evolves, particularly with the expansion of the RAG database, I expect it to become an even more powerful and cost-effective solution for securing large language models.

\section{CONCLUSION AND FUTURE WORK}
In this work, I presented RECAP, a novel method that leverages pre-generated adversarial tokens to improve the efficiency of adversarial attacks on large language models (LLMs). By bypassing the resource-intensive training processes required by traditional methods like GCG, RECAP provides a more accessible and cost-effective solution. My approach demonstrated comparable success rates while operating with minimal resources, making it particularly useful for organisations with limited computational capacity. Moreover, RECAP's applicability to black-box LLMs sets it apart from conventional adversarial methods, enabling testing in scenarios where model internals are inaccessible.

However, it is important to acknowledge the dual-use nature of adversarial research. While RECAP can be used to identify vulnerabilities in LLMs and strengthen them against malicious attacks, it also has the potential to be misused for generating harmful or unethical content. As researchers and developers, I emphasise the importance of using RECAP—and adversarial methods in general—for positive, constructive purposes. This tool should be employed to improve the robustness and safety of AI systems, helping organisations ensure their models are secure and aligned with ethical guidelines.

Looking forward, there are several avenues for future work. First, expanding the RAG database will be crucial in increasing the accuracy and success rate of adversarial attacks. A larger, more diverse token pool would allow RECAP to handle a wider range of adversarial scenarios and potentially outperform traditional methods without any training required. Additionally, I plan to refine my approach to further reduce training times by fine-tuning the integration of RECAP tokens into existing adversarial methods. By providing more effective token initialization, I aim to lower the computational burden even further while maintaining high success rates.

Another area of improvement lies in better understanding the inconsistencies I observed in some of the training scenarios. This could involve investigating the underlying factors that contribute to success and failure, such as token diversity, model architecture, and attack configurations.

In summary, RECAP holds great potential as a tool for secure, ethical, and cost-effective adversarial testing, with significant room for improvement in future iterations. Its continued development will contribute to the broader mission of building more secure, trustworthy AI systems for all.

\end{document}